%% file: LuXiao.tex
\documentclass[letterpaper, 10pt, conference]{ieeeconf}  % Comment this line out if you need a4paper
\IEEEoverridecommandlockouts                              % This command is only needed if 
\usepackage{booktabs}
\usepackage{graphicx}
\usepackage{array}
\usepackage{cite}
\usepackage{graphicx}
\usepackage{amsmath}
\usepackage{amssymb}
\usepackage{algorithm}
\usepackage[noend]{algorithmic}
\usepackage{flushend}
\usepackage[T1]{fontenc}
\usepackage{aecompl}
\usepackage{xurl}
\usepackage{hyperref}
\usepackage{ifthen}
\usepackage{caption}
\usepackage{multirow}
\newcommand{\Symbol}[1]{\ensuremath{\mathcal{#1}}}

\newcommand{\World}{\Symbol{W}}

\newcommand{\StateSpace}{\mathcal{X}}
\newcommand{\ActionSpace}{\Symbol{U}}

\usepackage{xcolor}
\definecolor{green}{RGB}{11,155,13}

\title{\bf Adaptive Dynamics Planning for Robot Navigation}

\author{Yuanjie Lu$^1$, Mingyang Mao$^2$, Tong Xu$^1$, Linji Wang$^1$, Xiaomin Lin$^2$, and Xuesu Xiao$^1$% <-this % stops a space
\thanks{$^1$Department of Computer Science, George Mason University, Virginia, USA.}%
\thanks{$^2$Department of Engineering Science, University of South Florida, Florida, USA}
}

\begin{document}

\maketitle

\begin{abstract} 
% background and challenge
Autonomous robot navigation systems often rely on hierarchical planning, where global planners compute collision-free paths without considering dynamics, and local planners enforce dynamics constraints to produce executable commands. This discontinuity in dynamics often leads to trajectory tracking failure in highly constrained environments.
% challenge
Recent approaches integrate dynamics within the entire planning process by gradually decreasing its fidelity, e.g., increasing integration steps and reducing collision checking resolution, for real-time planning efficiency. However, they assume that the fidelity of the dynamics should decrease according to a manually designed scheme. Such static settings fail to adapt to environmental complexity variations, resulting in computational overhead in simple environments or insufficient dynamics consideration in obstacle-rich scenarios. 
% motivation
To overcome this limitation, we propose Adaptive Dynamics Planning (ADP), a learning-augmented paradigm that uses reinforcement learning to dynamically adjust robot dynamics properties, enabling planners to adapt across diverse environments. We integrate ADP into three different planners and further design a standalone ADP-based navigation system, benchmarking them against other baselines. Experiments in both simulation and real-world tests show that ADP consistently improves navigation success, safety, and efficiency.

\end{abstract}

%\begin{IEEEkeywords}
%xxx
%\end{IEEEkeywords}

\input{Sections/Introduction}

\input{Sections/RelatedWork}

\input{Sections/ProblemFormulation}
\input{Sections/Methodology}
\input{Sections/Experiments}

\input{Sections/Conclusions}

\IEEEpeerreviewmaketitle

\bibliographystyle{IEEEtran}
\bibliography{ref.bib}

\end{document}

%% file: Sections/Introduction.tex
\section{Introduction}
\label{sec:Intro}

% Background and Importance)
% Significance of robot navigation in autonomous systems
Autonomous robot navigation is critical for robots to operate safely and efficiently in real-world environments. Planning collision-free trajectories while respecting dynamics constraints is particularly challenging in highly constrained environments, ranging from narrow indoor corridors to densely populated outdoor spaces. With the expanding deployment of robots in domains such as last-mile delivery, healthcare assistance, and industrial inspection, navigation systems must deliver both robustness and efficiency across varying environmental complexities.

\begin{figure}[!t]
\centering
\includegraphics[width=0.98\columnwidth]{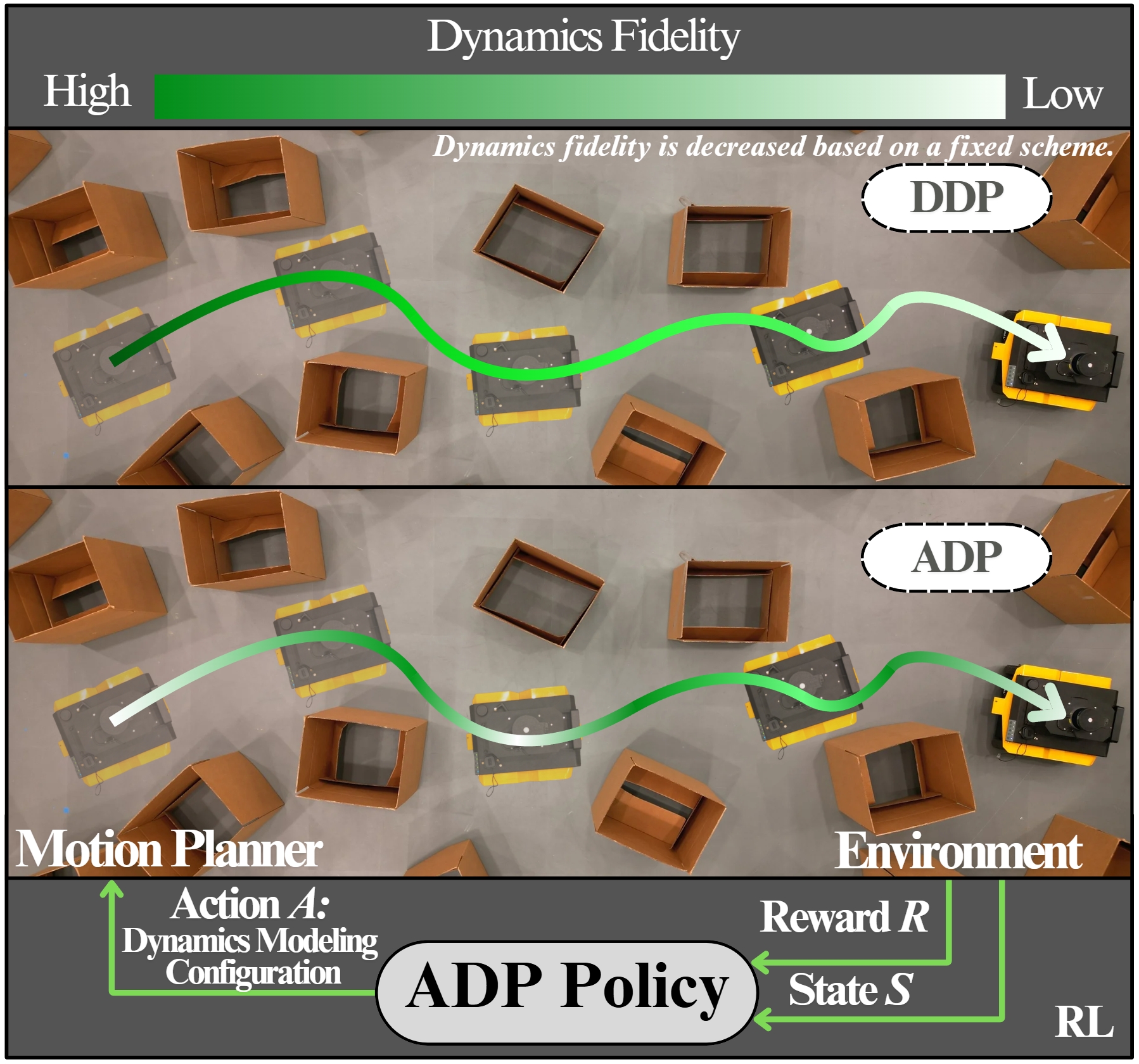}
\caption{Instead of using a fixed scheme to decrease dynamics fidelity as adopted by DDP~\cite{lu2025decremental} (top), ADP leverages Reinforcement Learning (RL) to adaptively adjust dynamics fidelity to balance modeling accuracy and computation efficiency during trajectory rollout (bottom).}
\label{fig:DDP}
\end{figure}

% Classical navigation planning 
Traditional navigation systems usually adopt a two-stage architecture that separates global path generation from local trajectory execution~\cite{tang2025path}. The global planning stage produces a geometrically feasible path over large configuration spaces by modeling the robot as a holonomic point mass, while the local planning stage manages real-time obstacle avoidance and motion execution under full dynamics constraints within limited horizons. This separation creates fundamental mismatches: global paths often ignore physical limits such as curvature and acceleration, forcing local planners to either heavily modify the path or fail in highly constrained environments~\cite{xiao2022motion}.

% Challenges in Existing Approaches
% 全局与局部规划器的解耦问题 Decoupled Treatment
% Inefficient resource allocation

Recent research has attempted to address this mismatch by proposing more integrated approaches to dynamics modeling. For example, guided sampling-based motion planning with dynamics~\cite{khanal2023guided} aims to generate trajectories that strictly satisfy the dynamics constraints along the entire path. Although theoretically appealing, such methods are computationally prohibitive for real-time navigation, given onboard resource limitations and the need for rapid replanning in dynamic environments. In contrast, Decremental Dynamics Planning (DDP)~\cite{lu2025decremental} tackles this computational challenge by starting with high-fidelity dynamics modeling at the early stages of trajectory rollout and gradually reducing model fidelity, thereby striking a balance between dynamics fidelity and computational efficiency. However, DDP's predetermined reduction schedules lack environmental awareness, resulting in computational overhead in simple environments where simplified models would suffice, or insufficient dynamics consideration in obstacle-rich environments that demand higher fidelity modeling.

To address these limitations, we propose Adaptive Dynamics Planning (ADP), a learning-augmented navigation paradigm that employs reinforcement learning to adapt dynamics modeling during the entire planning process. Unlike predetermined approaches like DDP that use fixed reduction of dynamics fidelity, ADP uses a learned agent as a meta-controller that dynamically adjusts dynamics model fidelity based on environmental observations. The agent learns to modulate various aspects of dynamics modeling—balancing computational efficiency with trajectory feasibility—while preserving the stability and interpretability of classical planning frameworks. This learned adaptation enables intelligent resource allocation that automatically matches modeling complexity to environmental demands.
To validate ADP's feasibility and effectiveness, we integrate it with three classical planners and compare it against fixed fidelity reduction approaches based on DDP. Experiments in the simulated BARN environments~\cite{perille2020benchmarking} and real-world deployment show consistent improvements in navigation success, safety, and efficiency.

%% file: Sections/RelatedWork.tex
\section{Related Work}
\label{sec:RelatedWork}
We review related work in autonomous robot navigation and addressing robot dynamics.

\subsection{Autonomous Robot Navigation}
Traditional robot navigation systems employ hierarchical architectures that separate global and local planning. Global planners, including grid-based~\cite{elfes1989using, dijkstra2022note} and sampling-based methods~\cite{lavalle2001rapidly, kavraki1996probabilistic}, compute coarse paths by modeling the robot as a holonomic point mass, neglecting dynamics for efficiency. Local planners, in contrast, incorporate dynamics constraints by forward-simulating motion over short horizons. Methods such as Dynamic Window Approach (DWA)~\cite{fox1997dynamic, fox2002dynamic} and Model Predictive Path Integral (MPPI) control~\cite{williams2017model, mohamed2022autonomous,  minavrik2024model} generate real-time commands while ensuring collision avoidance and path feasibility. While this decomposition appears to balance tractability with dynamics, it often yields paths that are geometrically valid but dynamically infeasible—for example, narrow passages requiring abrupt turns beyond the robot's steering limits or routes with acceleration demands that exceed actuator capabilities.

In addition to classical approaches, recent studies have explored learning-based navigation~\cite{xiao2022motion}. Early efforts focused on end-to-end approaches\cite{pfeiffer2017perception, kim2018end, amini2019variational} that directly map perceptual inputs to control commands, bypassing the global–local separation. While these methods mitigate coordination issues, they often lack the transparency and robustness required for safety-critical navigation. More recent work investigated hybrid approaches that combine learning with classical planning, aiming to retain the safety and interpretability of traditional methods while adding adaptability and flexibility. Such strategies enhance specific components of navigation pipelines, such as learning cost functions\cite{bui2022improving, lu2023leveraging, xiao2022learning}, predicting optimal parameters~\cite{xiao2022appl, wang2021apple, xu2021applr}, or modeling environmental conditions~\cite{lu2025multi, das2024motion}. Our proposed ADP method follows this hybrid paradigm but distinguishes itself by targeting robot dynamics: it employs a meta-control strategy to adapt dynamics modeling fidelity in real time to trade off computation efficiency within different classical planners.

\subsection{Robot Dynamics}
The computational burden of dynamics modeling poses a central challenge for real-time robot navigation. High-fidelity models that capture mass distribution, actuator limits, and environmental interactions enable accurate motion prediction but impose significant computational costs, whereas simplified models improve efficiency at the expense of physical realism and feasibility.

To balance this tradeoff, a wide spectrum of modeling approaches has been explored. Classical methods range from kinematic abstractions such as bicycle, Ackermann, and differential-drive models~\cite{plaku2018clearance, le2021multi, khanal2024learning} to analytical formulations incorporating inertial and actuator constraints. Learning-based approaches extend this spectrum with data-driven models that capture complex interactions for high-speed navigation~\cite{xiao2021learning, karnan2022vi,  pokhrel2024cahsor} and mobility on challenging off-road terrains~\cite{datar2024learning, datar2024terrain}. While high-fidelity models yield accurate state predictions for reliable action generation, their computational demands limit onboard applicability. To mitigate this, Decremental Dynamics Planning (DDP)~\cite{lu2025decremental} begins with high-fidelity modeling and gradually reduces fidelity during planning, trading accuracy for efficiency. However, DDP follows fixed reduction schedules without environmental awareness, which can result in inefficient resource allocation. To fully exploit computational resources and adapt to diverse environments, ADP leverages reinforcement learning to adapt dynamics fidelity in real time based on environmental observations, dynamically matching modeling complexity to task demands.

%% file: Sections/ProblemFormulation.tex
\section{Adaptive Dynamics Planning}
\label{sec:Problem}
We introduce ADP, which uses reinforcement learning to adjust dynamics modeling fidelity and efficiency during navigation adaptively. ADP learns when to use high-fidelity dynamics for accurate prediction and when to simplify dynamics models for computational efficiency. In this section, we first formulate the motion planning problem, then define ADP under the Markov Decision Process (MDP) framework, and finally discuss the learning algorithm.

\subsection{Motion Planning Problem Formulation}
We consider a mobile robot operating in a world $\World$ with state space $\StateSpace = \StateSpace_{\text{free}} \cup \StateSpace_{\text{obs}}$, where $\StateSpace_{\text{free}}$ and $\StateSpace_{\text{obs}}$ denote free space and obstacle regions, respectively. The robot's control space is $\ActionSpace$ and the robot's motion follows a dynamics model:
\begin{equation}
s_{t+1} = f(s_t, u_t; \phi_t), \quad s_t \in \StateSpace, \, u_t \in \ActionSpace,
\label{eqn::motion}
\end{equation}
where $\phi_t$ specifies the dynamics modeling configuration at time step $t$. In general, $\phi_t$ can be categorized into two types: (i) physical constants that characterize the robot's mechanical properties, such as mass, inertia, wheelbase, and axle length; and (ii) computational parameters that regulate the trade-off between modeling fidelity and computational efficiency, such as integration interval, collision-checking resolution, or in the case of learned dynamics models based on neural networks, (number of) weights and biases.

We instantiate a differential-drive robot as an example. The state $s_t = (x, y, \psi)$ represents the robot’s planar pose in a fixed global coordinate system, 
where $(x,y)$ denote translations along the $\mathbf{x}$- and $\mathbf{y}$-axes, and $\psi$ denotes the yaw angle around the $\mathbf{z}$-axis at time step $t$. The control input $u_t$ is determined by the differential-drive robot's kinematic constraints, specifically $u_t = (v_t, w_t)$, where $v_t$ denotes the linear velocity and $w_t$ represents the angular velocity. The dynamics configuration $\phi_t$ can vary over time according to a selection strategy that maps the current state and observations to appropriate modeling parameters. Different approaches determine this strategy in different ways. 
Traditional navigation systems employ static thresholds (e.g., a time threshold $\tau$ or a distance threshold $d$) to switch between global and local planning phases, maintaining fixed parameter sets throughout navigation. In contrast, Decremental Dynamics Planning (DDP) adopts hand-crafted schedules that gradually reduce modeling fidelity along the rollout by adjusting integration interval and collision-checking resolution over time.

In summary, given a start state $s_0 \in \StateSpace_{\text{free}}$ and a goal state $s_g \in \StateSpace_{\text{free}}$, the motion planning problem that considers dynamics is to find a sequence of control inputs $\{u_t\}_{t=0}^{T-1}$ together with a sequence of dynamics modeling configurations $\{\phi_t\}_{t=0}^{T-1}$ to generate trajectories that satisfy $s_t \in \StateSpace_\text{free}, \forall t \in \{t\}^{T}_{t=0}$ and ensure dynamic feasibility. 

\subsection{Adaptive Dynamics Planning Problem Formulation}
We formulate ADP as a MDP operating within a meta-environment $\mathcal{E}$ that encompasses both the obstacle-occupied world $\mathcal{W}$ and the underlying motion planner $p$. The MDP is defined by the tuple $(\mathcal{S}, \mathcal{A}, \mathcal{T}, \mathcal{R}, \gamma)$, where the agent's objective is to adaptively select dynamics modeling configurations based on environmental conditions. 

The state space $\mathcal{S}$ at time step $t$ is defined as $s_t = (o_t, \phi_{t-1}, g_t)$, where $o_t$ represents the current sensor observations (e.g., laser scans), $\phi_{t-1}$ denotes the previous dynamics modeling configuration, and $g_t$ encodes goal-related information, including the relative distance and angle to the local goal. The action space $\mathcal{A} = \Phi$ defines the feasible parameter space for dynamics modeling configurations, encompassing feasible ranges for integration intervals, trajectory rollout horizons, collision-checking densities, velocity/acceleration limits, and learned dynamics model parameters. Each configuration $\phi \in \Phi$ represents specific parameter values within these ranges. The transition function $\mathcal{T}: \mathcal{S} \times \mathcal{A} \to \mathcal{S}$ operates within $\mathcal{E}$ as follows: when the agent selects action $a_t \in \mathcal{A}$, determining configuration $\phi_t = a_t$, the motion planner $p$ rolls out trajectories based on a dynamics model that uses this configuration to select best control commands $u_t$, which are executed in world $\mathcal{W}$. This results in the next state $s_{t+1} = (o_{t+1}, \phi_t, g_{t+1})$, where $s_{t+1} \sim \mathcal{T}(\cdot|s_t, \phi_t)$ includes new observations and goal information. To guide the learning process, the reward function $\mathcal{R}: \mathcal{S} \times \mathcal{A} \to \mathbb{R}$ evaluates the trajectory performance resulting from applying dynamics configuration $\phi_t$ over the planning horizon, encouraging configurations that achieve necessarily accurate dynamics prediction while maintaining computational efficiency.

Finally, the objective of ADP is to learn an optimal policy $\pi^*: \mathcal{S} \to \mathcal{A}$ that maximizes the expected cumulative reward over time:
\begin{equation}
\max_{\pi} J^{\pi} = \mathbb{E}_{s_0,\phi_t\sim\pi(s_t),s_{t+1}\sim\mathcal{T}(s_t,\phi_t)} \left[ \sum_{t=0}^{T} \gamma^t r_t \right],
\nonumber
\end{equation}
where $\gamma \in [0,1]$ is the discount factor.

\begin{figure*}[!t]
  \centering  
  \includegraphics[width=\textwidth]{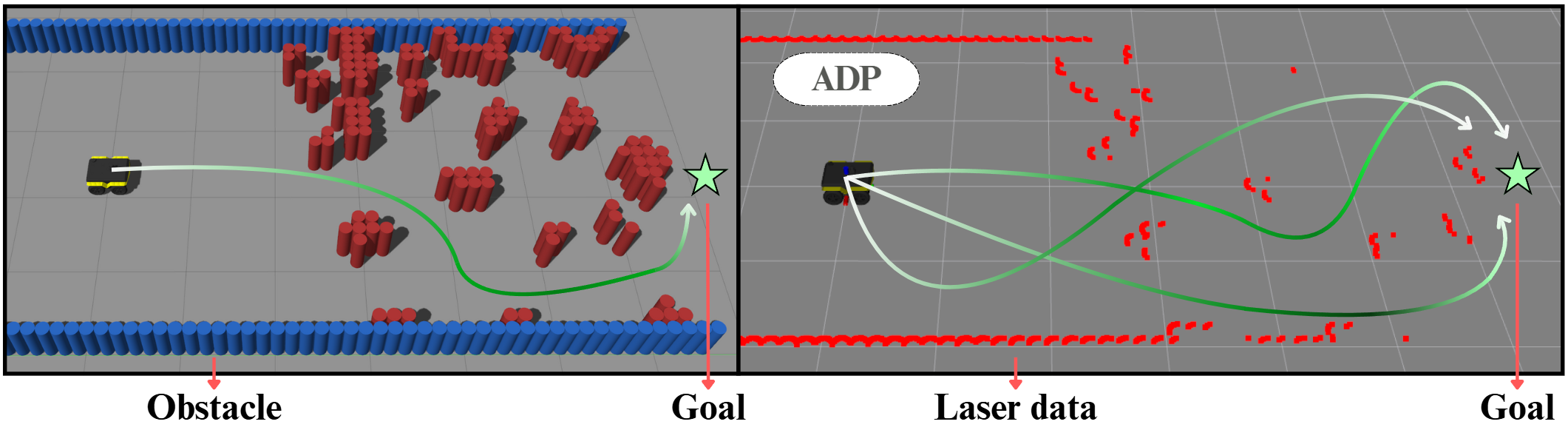}\\[-2mm]
  \caption{Example of the ADP Navigation System Operating in a BARN Environment. Left: Gazebo Visualization. Right: RViz Visualization. Lighter and deeper green colors indicate lower and higher fidelity in dynamics modeling, respectively.}
  \label{fig:simulation}
\end{figure*}

\subsection{State, Action, and Reward Specification}
\textbf{State Space:} Our specific state space includes four components: (i) A 720-dimensional laser scan with a $270^\circ$ field of view, normalized to the range $[-1,1]$; (ii) The previous dynamics configuration from the last time step; (iii) The robot’s current linear and angular velocities; (iv) Goal-related information, including the relative distance and bearing to both a local goal and the global goal. These components are concatenated into a fixed-size state vector that serves as the state representation for learning, consisting of 726 dimensions plus the dimensionality of the previous action space.

\textbf{Action Space:} The action space defines parameters for adaptive dynamics modeling. We design multiple strategies: First, fixed parameter selection, where actions directly specify values for integration intervals and collision checking density; Second, a decremental dynamics strategy where actions specify total rollout time $T$, discretization steps $N$ (determining how many time intervals the rollout horizon is divided into), temporal distribution parameter $p$, and blending coefficient $\alpha$. This strategy focuses on adaptive integration intervals and does not modify collision checking density. The time intervals are computed as $\Delta t_i = t_i - t_{i-1}$ where $t_i = \alpha \cdot \frac{i}{N} \cdot T + (1-\alpha) \cdot (\frac{i}{N})^p \cdot T$; Third, an incremental dynamics variant that reverses the computed interval sequence from decremental dynamics function. We hypothesize that the counterintuitive incremental dynamics approach would not achieve good results. Actions are normalized to $[-1,1]$ during training and mapped to feasible parameter ranges during execution.

\textbf{Reward Design:} The reward function encourages effective navigation through four components: (1) progress reward $r_\text{progress}$ measuring distance reduction toward the goal, (2) collision penalty $r_\text{collision}$ enforcing safety constraints, (3) time penalty $r_\text{time}$ promoting efficient task completion, and (4) obstacle avoidance reward $r_\text{obstacle}$ encouraging safe proximity to environmental obstacles. Specifically, the obstacle avoidance component identifies the 10 nearest obstacles from laser scan data and applies a quadratic penalty $(1 - d/0.05)^2$ when the robot-obstacle distance $d < 0.05$m, encouraging the robot to maintain safe distances from environmental obstacles. The total reward is $r_t = r_\text{progress} + r_\text{collision} + r_\text{time} + r_\text{obstacle}$. This formulation guides the agent to learn dynamics parameter selection policies that optimize navigation performance by adapting dynamics modeling to environmental demands.

\subsection{Reinforcement Learning Algorithm}
ADP requires a reinforcement learning algorithm that handles continuous parameter spaces and demonstrates high sample efficiency in navigation scenarios. TD3~\cite{fujimoto2018addressing} (Twin Delayed Deep Deterministic Policy Gradient) meets these requirements as an off-policy actor-critic method designed for continuous control. 
The algorithm maintains a policy network and dual Q-function approximators. 
The dual Q-network~\cite{hasselt2010double} design addresses overestimation bias through double learning, where each network updates via the Bellman residual objective. 
Training stability is enhanced through delayed policy updates and target value clipping. 
% The algorithm maintains a policy network $\pi^{\xi}_p$ and dual Q-function approximators $Q^{\zeta_1}_p$ and $Q^{\zeta_2}_p$, parameterized by $\xi$ and $\zeta$ respectively. Policy updates follow the deterministic gradient rule~\cite{silver2014deterministic}:
% \begin{equation}
% \nabla_{\xi}J^{\pi^{\xi}_p} = \mathbb{E}_{s\sim\pi^{\xi}_p}\left[\nabla_a Q(s, a)|_{a=\pi^{\xi}_p} \nabla_{\pi^{\xi}_p}(s)\right]
% \end{equation}
% The dual Q-network~\cite{hasselt2010double} design addresses overestimation bias through double learning, where each network updates via the Bellman residual objective:
% \begin{equation}
% \mathbb{E}_{(s,a,r,s')\sim\mathcal{E}}\left[||Q^{\zeta}_p(s, a) - r - \gamma \max_{a'} Q^{\zeta}_p(s', a')||^2\right]
% \end{equation} 
% Training stability is enhanced through delayed policy updates and target value clipping using $\min(Q^{\zeta_1}_p(s, a), Q^{\zeta_2}_p(s, a))$. 
Given the computational demands and variance inherent in navigation simulation, TD3's robustness and sample efficiency make it well-suited for learning dynamics parameter policies. To improve sample efficiency, we employ distributed training using SLURM workload manager with multiple parallel actors collecting experience across different simulation environments. Each actor collects data from at most two episodes before entering an idle state, waiting for the next assignment cycle. Data is aggregated in a centralized replay buffer for a single learner process, accelerating training while providing environmental diversity for robust policy learning.

%% file: Sections/Methodology.tex
\subsection{ADP-Augmented Planners}
Based on the ADP design, we integrate adaptive dynamics modeling into three sampling-based motion planners: DWA~\cite{fox1997dynamic}, MPPI~\cite{williams2017model}, and Log-MPPI~\cite{mohamed2022autonomous}: 
DWA samples velocity commands within dynamic constraints and simulates resulting trajectories; MPPI optimizes trajectories through importance sampling of randomly perturbed control sequences, computing optimal actions as weighted averages; Log-MPPI applies exponential transformation to trajectory costs, combining stochastic sampling with model predictive control for robust optimization.

All three planners use the same ADP temporal discretization strategy, where the learned policies are trained separately and adaptively adjust integration intervals based on environmental conditions. This replaces fixed parameters in vanilla implementations and hand-crafted schedules in DDP. Apart from the dynamics modeling precision selection, all other components remain identical to DDP counterparts, ensuring fair comparison. Each ADP-augmented planner operates independently without recovery behaviors, relying solely on learned adaptive dynamics for robust navigation.

\subsection{Standalone ADP-based navigation system}
The ADP navigation system incorporates the same operational modes as DDP, including high-speed navigation, precision maneuvering, and recovery behaviors. The system transitions between these modes based on environmental feedback, maintaining the proven mode-switching framework established by DDP. However, while DDP relies on hand-crafted schedules for dynamics parameter selection, ADP employs learned policies to adaptively configure temporal discretization. In open areas, ADP automatically selects coarse integration intervals with fewer discretization steps for computational efficiency, while complex environments trigger fine-grained integration intervals with higher step counts for enhanced precision, with these decisions driven by reinforcement learning rather than predetermined rules. 

At each planning iteration, the motion planner receives these adaptive parameters and generates trajectory samples using the variable temporal resolution. The learned discretization strategy directly affects both the prediction accuracy of individual trajectories and the overall computational cost of the planning process. Trajectory evaluation employs a comprehensive cost function incorporating goal proximity, obstacle clearance, path efficiency, and motion smoothness. After evaluating all candidates, the system selects the $N=10$ lowest-cost collision-free trajectories and computes control commands through weighted averaging based on trajectory costs. This approach ensures that the final control decisions reflect both the quality of individual trajectories and the reliability of their underlying dynamics modeling.

\begin{table*}
    \centering
    \caption{Performance Comparison of Adaptive and Decremental Dynamics Planning: Experimental Evaluation on 225 Testing Environments from BARN Challenge}
    \setlength{\tabcolsep}{5.85pt}
    \begin{tabular}{lccccccccccccccc}
        \toprule
        \textbf{Method} & \multicolumn{3}{c}{\textbf{Task Success (\%)} $\uparrow$} & \multicolumn{3}{c}{\textbf{Avg. Time (s)} $\downarrow$} & \multicolumn{3}{c}{\textbf{Avg. Score} $\uparrow$} & \multicolumn{3}{c}{\textbf{Avg. Collision (\%)} $\downarrow$} & \multicolumn{3}{c}{\textbf{Avg. Timeout (\%)} $\downarrow$} \\
        \cmidrule(lr){2-4} \cmidrule(lr){5-7} \cmidrule(lr){8-10} \cmidrule(lr){11-13} \cmidrule(lr){14-16}
        & 1.0 & 1.5 & 2.0 & 1.0 & 1.5 & 2.0 & 1.0 & 1.5 & 2.0 & 1.0 & 1.5 & 2.0 & 1.0 & 1.5 & 2.0 \\
        \midrule 
        DWA-DDP  &   93.74  &  86.46 & 73.94  & 12.47 &  13.10 & 17.67 & 0.469  &  0.432  & 0.370 & 06.26   &  13.54  &  26.06 &  \textbf{0.00}   &  \textbf{0.00}  &  \textbf{0.00} \\
        DWA-ADP &   97.03  &  94.98 & 82.20  & \textbf{10.47}  & 08.62  & 11.82 & 0.485  &  0.475 &  0.411 &  02.97   &  05.02  & 17.80  &  \textbf{0.00}   & \textbf{0.00} & \textbf{0.00} \\
        \midrule
        MPPI-DDP &  91.45   &  90.54 &  82.39 & 14.46 & 12.11  & 14.31  & 0.457  & 0.453  &  0.412 &  08.41   &  09.32  &  17.61 &  0.14   &  0.15  & \textbf{0.00}  \\
        MPPI-ADP &  96.70   & 94.13  & 93.28  & 11.56 & \textbf{08.22}  & 10.28 &  0.482 & 0.461  &  0.466 & 01.02    &  01.01  &  \textbf{00.33} &   2.28  &  4.86  &  6.40 \\
        \midrule
        Log-MPPI-DDP &  61.24  &  53.86 & 50.41  & 26.79 & 27.54  & 31.15 &  0.294 & 0.269  &  0.240 &  34.74   &  45.71  & 46.81  &   4.02  &  0.44  & 2.78  \\
        Log-MPPI-ADP &   65.85  & 57.82  & 54.38  & 22.46 & 21.74  & 25.14 & 0.320  & 0.298  & 0.264 & 31.31  & 42.11   & 43.45  &  2.84   &  0.07  &  2.06  \\
        \midrule
        DDP & 95.38  &  96.61 &  93.05 & 14.28 &  11.63 & 11.32 & 0.473  & 0.483  & 0.465 &   01.02  &  \textbf{00.00}  &  05.03 &  3.60   &  3.39  & 1.92 \\ 
        ADP &   \textbf{99.61}  & \textbf{99.22}  & \textbf{99.11}  & 12.85 & 10.22  & \textbf{08.66} &  \textbf{0.491} & \textbf{0.494}  & \textbf{0.494} &  \textbf{00.33}   & 00.59   & 00.60  &  0.07   &  0.20  & 0.29  \\
        \bottomrule

    \end{tabular}
    \label{tab:DDP}

    \vspace{4pt}
    \parbox{\textwidth}{\raggedright Each environment is tested 20 times, with the 5 best and 5 worst results excluded. Values 1.0, 1.5, and 2.0 denote maximum linear velocity (m/s). Bold indicates the best result.}

\end{table*}

\begin{table}[t]
    \centering
    \renewcommand{\arraystretch}{1.2}
    \caption{Ablation Studies on ADP Performance}
    \label{tab:ablation_combined}

    \setlength{\tabcolsep}{5.5pt}
    \begin{tabular}{lccc}
        \toprule
        \multicolumn{4}{c}{\textbf{Impact of Different ADP Models}} \\
        \midrule
        \textbf{Method} & \textbf{Success (\%)} $\uparrow$ & \textbf{Avg. Time (s)} $\downarrow$ & \textbf{Avg. Score} $\uparrow$  \\
        \midrule
        ADP-Unc & 99.20 & 10.55 & 0.491   \\
        ADP-Dec & \textbf{99.22} & \textbf{10.22} & \textbf{0.494}   \\
        ADP-Inc  & 96.85 & 11.87 & 0.480  \\
        \bottomrule
    \end{tabular}
    
    \vspace{10pt}

    \begin{tabular}{lccc}
        \toprule
        \multicolumn{4}{c}{\textbf{Impact of Training Environment Quantity}} \\
        \midrule
        \textbf{Method} & \textbf{Success (\%)} $\uparrow$ & \textbf{Avg. Time (s)} $\downarrow$ & \textbf{Avg. Score} $\uparrow$  \\
        \midrule
        ADP-25 & 96.87 & 11.16 & 0.484   \\
        ADP-75 & \textbf{99.22} & \textbf{10.22} & \textbf{0.494}   \\
        ADP-125  & 99.11 & 10.79 & \textbf{0.494}  \\
        ADP-175  & 98.92 & 10.98 & 0.491  \\
        \bottomrule
    \end{tabular}
    
    \vspace{4pt}
    \parbox{\textwidth}{\raggedright 
    Maximum linear velocity: 1.5 m/s.}
\end{table}

%% file: Sections/Experiments.tex
\section{Experiments}

\label{sec:ExpResults}
In this section, we validate ADP’s capability to adapt dynamics modeling across diverse environments without manual tuning or fixed reduction schedules. ADP is integrated into three classical planners and evaluated in direct comparison with their DDP-augmented counterparts. Experiments are conducted in both the BARN Challenge benchmark~\cite{xiao2022autonomous, xiao2023autonomous, xiao2024autonomous} and real-world deployments using a Clearpath Jackal robot.

\subsection{Experimental Setup}
\subsubsection{Baseline Methods}
While DDP has shown improvement when augmenting classical approaches~\cite{lu2025decremental} like DWA~\cite{fox1997dynamic}, MPPI~\cite{williams2017model}, and Log-MPPI~\cite{mohamed2022autonomous}, we evaluate the DDP-augmented DWA-DDP, MPPI-DDP, and Log-MPPI-DDP against our ADP variants, i.e., DWA-ADP, MPPI-ADP, and Log-MPPI-ADP, to assess learned versus predetermined dynamics adaptation. Additionally, we compare integrated DDP and ADP navigation systems that incorporate recovery behaviors, a capability not provided by the three standalone planners. All DWA and MPPI variants directly process laser scans with goal progress, obstacle avoidance, and motion smoothness objectives, while Log-MPPI variants employ costmap-based planning. All DDP-based planners (except Log-MPPI variants) operate with a 2-second trajectory rollout horizon discretized into 20 time steps with temporal resolution $\Delta t_i = \left(\frac{i}{N}\right)^p \cdot T_\text{total} - \left(\frac{i-1}{N}\right)^p \cdot T_\text{total}$, where $N=20$, $T_\text{total}=2$ seconds, and $p=1.7$. DWA variants use 20×20 linear and angular velocity grid sampling, MPPI variants generate 550 trajectory samples with 8-thread parallel processing, while Log-MPPI variants employ default parameters from the public implementation.

% \begin{figure}[t]
% \centering
% \includegraphics[width=0.95\columnwidth]{figures/environment1.pdf}
% \caption{Two Physical Test Environments.}
% \label{fig:box}
% \end{figure}

\subsubsection{Experimental Environments}
\begin{figure}[t]
\centering
\includegraphics[width=0.98\columnwidth]{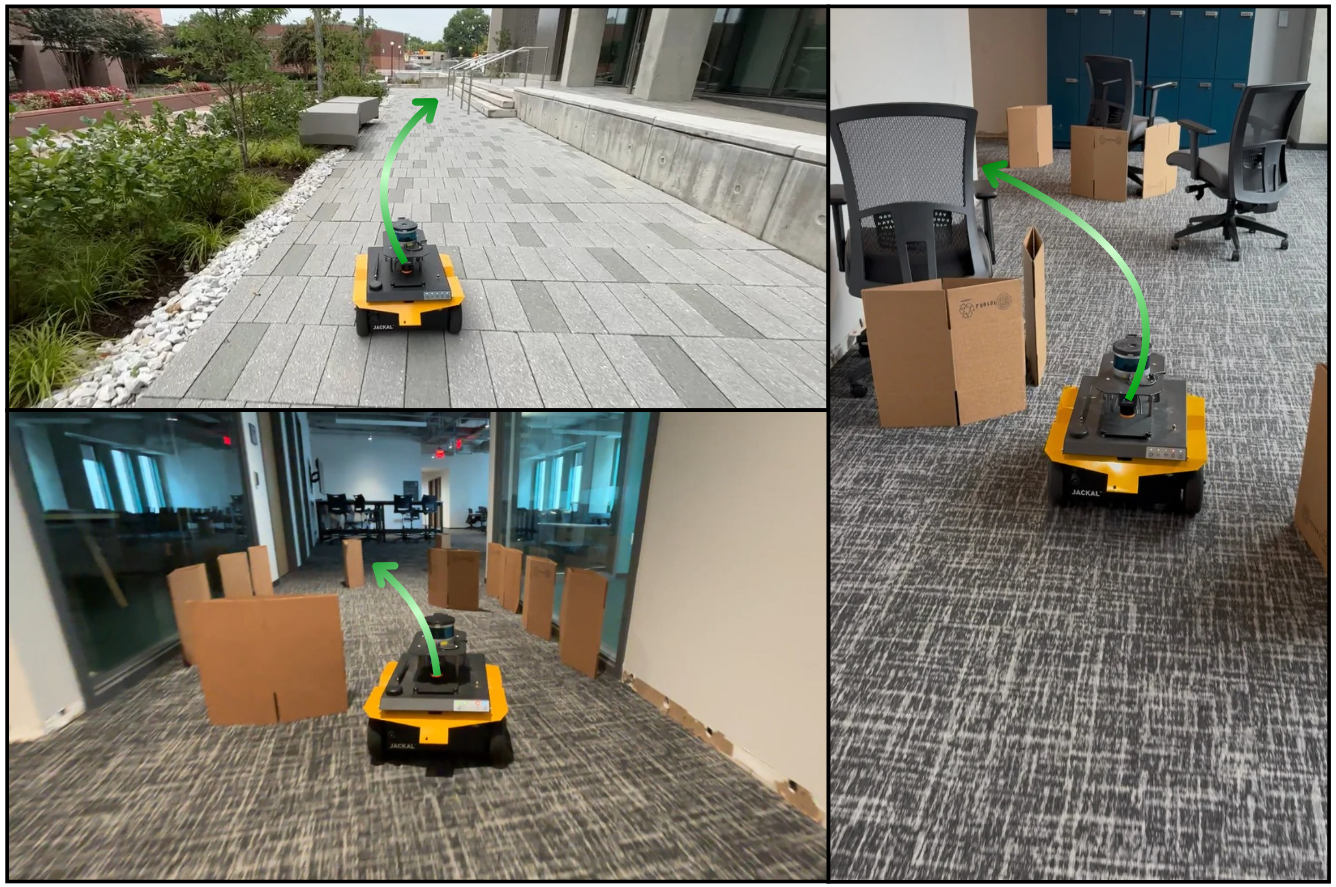}
\caption{Real-World Cluttered Environment Experiments.}
\label{fig:physical2}
\end{figure}

Our evaluation encompasses both simulated and real-world scenarios. The BARN benchmark~\cite{perille2020benchmarking} provides 300 environments with diverse obstacle configurations generated via cellular automata (Fig.~\ref{fig:simulation}). We randomly select 75 training environments from the 150 most challenging ones, with the remaining 225 environments serving as the test set. Additional evaluation is conducted on 50 unpublished environments from the 2025 BARN Challenge~\cite{xiaoautonomous} to compare against state-of-the-art navigation methods. Physical experiments are conducted in indoor test courses, where the robot starts from the bottom-left corner and navigates to a star-shaped goal in the upper-right corner (Fig.~\ref{fig:physical}). Additionally, ADP is deployed in natural cluttered indoor and outdoor environments (Fig.~\ref{fig:physical2}).
% Our evaluation encompasses both simulated and real-world scenarios. The BARN benchmark~\cite{perille2020benchmarking} provides 300 environments with diverse obstacle configurations generated via cellular automata. We randomly select 75 training environments from the 150 most challenging ones, with the remaining 225 environments serving as the test set. Additional evaluation is conducted on 50 unpublished environments from the 2025 BARN Challenge~\cite{xiaoautonomous} to compare against state-of-the-art navigation methods. Physical experiments are conducted in indoor test courses with defined start and goal positions, as well as natural cluttered spaces, to assess real-world navigation capabilities. Fig.~\ref{fig:simulation} and Fig.~\ref{fig:physical} show representative simulated and physical test environments.

\subsubsection{Training Configuration}
ADP agents use TD3 with actor and critic networks. The actor employs a single fully-connected layer after feature extraction with tanh activation for bounded outputs, while critics use 3-layer networks (512-256-1 units) with ReLU activations. Training occurs on a SLURM cluster, with each training environment generating two episodes before entering standby mode to prevent data imbalance and ensure uniform sampling across different environments. The system initializes from the DDP dynamics schedule and learns environment-aware adaptation using experience replay (200K buffer) and n-step returns (n=6) for improved sample efficiency.

\subsubsection{Hardware Configuration}
Physical experiments are conducted on a Clearpath Jackal robotic platform equipped with a Hokuyo LiDAR sensor providing 720-dimensional laser scans over a 270° field of view. Simulation experiments run on a computing platform featuring an AMD Ryzen 9 5900X processor (3.7 GHz) under Ubuntu 20.04 with ROS Noetic. All implementations are developed in C++ using g++ 9.4.0 for optimal real-time performance.

\subsubsection{Evaluation Metrics}
In simulation, we adopt the BARN Challenge scoring metric to quantify navigation performance. Each environment is assigned a score based on task completion and traversal efficiency:
\begin{equation}
s_i = 1^{\text{success}} \times \frac{\text{OT}_i}{\text{clip}(\text{AT}_i, 2\text{OT}_i, 8\text{OT}_i)}
\nonumber
\end{equation}
where $\text{OT}_i$ and $\text{AT}_i$ denote the optimal and actual traversal times respectively with $1^{\text{success}}$ as an indicator function for traversal success. The clipping bounds normalize execution time, ensuring scores remain within a consistent range~\cite{xiao2022autonomous, xiao2023autonomous, xiao2024autonomous}. To compute the average traversal time across all trials, the traversal time of failed trials is set to 50 seconds. For each environment, we conduct 20 runs and report results after removing the top five and bottom five performance trials to ensure statistical robustness. For real-world experiments, performance is evaluated using three primary metrics: success rate, completion progress, and traversal time.

\begin{table}[t]
    \centering
    \renewcommand{\arraystretch}{1.2}
    \setlength{\tabcolsep}{5.5pt}
    \caption{Performance Comparison on BARN Challenge Environments}
    \label{tab:barn_combined}
    
    % 第一个表格
    \begin{tabular}{lccc}
        \toprule
        \multicolumn{4}{c}{\textbf{50 Most Challenging BARN Test Environments}} \\
        \midrule
        \textbf{Method} & \textbf{Success (\%)} $\uparrow$ & \textbf{Avg. Time (s)} $\downarrow$ & \textbf{Avg. Score} $\uparrow$  \\
        \midrule
        DWA-DDP  & 44.00 & 31.74 & 0.220   \\
        DWA-ADP & 80.00 & 14.81 & 0.399 \\
        \midrule
        MPPI-DDP  & 50.67 & 29.90 & 0.253  \\
        MPPI-ADP  & 80.30 &  \textbf{13.36} &  0.401  \\
        \midrule
        Log-MPPI-DDP  & 02.67 & 48.88 & 0.013 \\
        Log-MPPI-ADP  & 08.51 & 41.87 & 0.085  \\
        \midrule
        DDP  & 79.63 & 21.33 & 0.397  \\
        ADP  & \textbf{95.80} & 14.92 & \textbf{0.470}  \\
        \bottomrule
    \end{tabular}
    
    \vspace{10pt}
    
    % 第二个表格
    \begin{tabular}{lccc}
        \toprule
        \multicolumn{4}{c}{\textbf{2025 BARN Challenge Leaderboard (50 Unpublished Environments)}} \\
        \midrule
        \textbf{Method} & \textbf{Success (\%)} $\uparrow$ & \textbf{Avg. Time (s)} $\downarrow$ & \textbf{Avg. Score} $\uparrow$ \\
        \midrule
        INVENTEC  & 98.20 & 14.13 & 0.4206  \\
        KUL+FM  & \textbf{99.60} & 12.32 & 0.4641  \\
        AIMS & 96.00 & 9.70 & 0.4723 \\
        LiCS-KI~\cite{damanik2024lics}   & 95.40 & \textbf{7.55} & 0.4762 \\
        DDP   & 99.00 & 7.67 & 0.4873 \\
        FSMT   & 98.20 & 8.65 & 0.4878 \\
        \midrule
        ADP   & 99.39 & 10.23 & \textbf{0.4940} \\
        \bottomrule
    \end{tabular}
    
    \vspace{4pt}
    \parbox{\textwidth}{\raggedright All approaches are evaluated with a maximum linear velocity of 1.5 m/s.}
\end{table}

% \begin{figure}[t]
% \centering
% \includegraphics[width=0.98\columnwidth]{figures/environment.pdf}
% \caption{Experiments in Real-World, Natural Cluttered Spaces.}
% \label{fig:chair}
% \end{figure}

\subsection{Ablation Study}
\subsubsection{Dynamics Modeling Strategies}
The first table in Table~\ref{tab:ablation_combined} compares three ADP dynamics modeling strategies: ADP-Unc (unconstrained) allows flexible dynamics fidelity adjustment across all time steps without predefined patterns, ADP-Dec (decremental) reduces dynamics fidelity over the planning horizon, and ADP-Inc (incremental) progressively increases dynamics fidelity. ADP-Dec achieves the best overall performance, demonstrating a superior success rate, time efficiency, and navigation score compared to the other variants. ADP-Inc shows notably lower performance across all metrics, suggesting that increasing dynamics fidelity over time is less effective for navigation tasks. These results validate the effectiveness of decremental scheduling proposed by DDP, which necessitates adaptation in contrast to a fixed schedule. 

\subsubsection{Number of Training Environments}
The second table in Table~\ref{tab:ablation_combined} examines the effects of training environment quantity on ADP performance. ADP-25 underperforms due to insufficient training diversity, while ADP-75 achieves better results across all metrics. Interestingly, ADP-125 maintains similar performance to ADP-75, but ADP-175 shows slight degradation. 

Based on these findings, we use ADP-Dec trained on 75 environments as our default configuration, which achieves the best performance while avoiding overfitting to the training distribution.

\subsection{Simulated Experiments}
\subsubsection{Standard BARN Environments}
Table~\ref{tab:DDP} compares ADP and DDP variants across 225 test BARN environments at three velocity settings, with navigation times including both successful and failed attempts. ADP-enhanced planners consistently surpass their DDP counterparts, achieving higher success rates, fewer collisions, and faster navigation across DWA, MPPI, and Log-MPPI. The standalone ADP system works particularly well, sustaining near-perfect navigation even at maximum velocity, where fixed-schedule methods typically degrade. These findings highlight the effectiveness of adaptive dynamics modeling, which delivers both safety and efficiency. Log-MPPI exhibits inherently lower performance due to the difficulty in maintaining accurate costmaps during agile maneuvers in highly constrained spaces, though it still demonstrates improvement with ADP. These results suggest that environment-aware dynamics adaptation can achieve improvements within various planning paradigms.

\subsubsection{Challenging Environments}
The top table in Table~\ref{tab:barn_combined} reports the performance across the 50 most challenging BARN test environments. The benefits of ADP become even more pronounced in these demanding scenarios: adaptive modeling enables planners to remain robust under extreme constraints, while DDP counterparts suffer sharp performance drops. ADP-enhanced DWA and MPPI maintain strong reliability, and the standalone ADP system continues to succeed in scenarios where other methods frequently fail. Even Log-MPPI, though inherently limited by its costmap-based structure, gains measurable improvements from ADP integration. These results confirm that adaptive dynamics modeling is especially critical in highly constrained settings.

\subsubsection{Unpublished environments}
The bottom table in Table~\ref{tab:barn_combined} compares ADP against top-performing methods from the 2025 BARN Challenge using 50 unpublished test environments. ADP achieves the strongest overall performance, surpassing prior systems, including DDP. This highlights ADP’s strong generalization ability beyond its training distribution and underscores its potential for broader deployment in diverse navigation scenarios.

\subsection{Physical Experiments}

\begin{figure}[t]
\centering
\includegraphics[width=0.98\columnwidth]{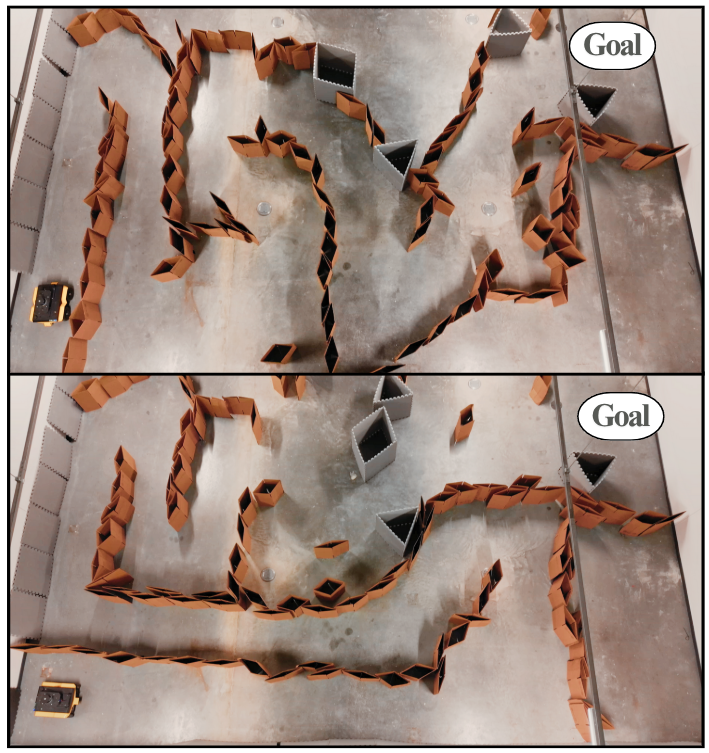}
\caption{Two Physical Test Environments.}
\label{fig:physical}
\end{figure}

% For real-world experiments (Fig.\ref{fig:box}), we design two types of complex environments and test five navigation methods: DWA, DWA-DDP, MPPI, MPPI-DDP, and DDP. Due to the poor performance of Log-MPPI in previous experiments and safety considerations, we exclude it from physical testing. Each method is evaluated five times in each environment. For successful task completions, we calculate the average time; for failed attempts, we measure the average navigation progress. As shown in Table \ref{tab:real_world}, our DDP-augmented methods demonstrate superior ability to make significant progress even in failed attempts. For the scenarios in Fig.~\ref{fig:chair}, we conduct experiments exclusively with our DDP method, which successfully navigates through both natural environments.

As shown in Fig.~\ref{fig:physical}, we evaluate three groups of navigation methods across two complex physical environments for real-world experiments: DWA-DDP vs DWA-ADP, MPPI-DDP vs MPPI-ADP, and the standalone systems DDP vs ADP. Due to Log-MPPI's poor performance in simulation experiments and safety considerations for the physical testing, we exclude it from real-world evaluation. Each method is tested four times in each environment. For successful completions, we record the average traversal time; for failed attempts, we set a uniform timeout of 2 minutes and measure navigation progress. Table~\ref{tab:real_world} shows that ADP-augmented methods consistently outperform their DDP counterparts across all metrics. Both standalone systems (DDP and ADP) achieve perfect success rates, with ADP demonstrating significantly faster completion times. ADP shows notable improvements in both success rates and progress completion compared to their respective baselines, validating the effectiveness of learned adaptive dynamics over hand-crafted scheduling in real-world scenarios.

\begin{table}[t]
    \centering
    \renewcommand{\arraystretch}{1.2} % Increase row height
    \caption{Performance on Two Physical Test Environments}
    
    \begin{tabular}{lcccc}
        \toprule
        \textbf{Method} & \textbf{Success} $\uparrow$ & \textbf{Avg. Progress} (\%) $\uparrow$ & \textbf{Avg. Time (s)} $\downarrow$  \\
        \midrule
        DWA-DDP  & 0/8 & 67.52 $\pm$ 4.67 & 120.00 $\pm$ 0.00  \\ 
        DWA-ADP & 2/8 & 77.21 $\pm$ 4.35 & 101.57 $\pm$ 4.31 \\
        \midrule
        MPPI-DDP   & 3/8 & 74.10 $\pm$ 5.72 & 114.87  $\pm$ 5.21 \\
        MPPI-ADP  & 5/8 &  89.50 $\pm$ 1.82 & 95.75 $\pm$ 4.57 \\
        \midrule
        DDP  & \textbf{8/8} & \textbf{100.00 $\pm$ 0.00} & 90.24 $\pm$ 4.73 \\
        ADP  & \textbf{8/8} & \textbf{100.00 $\pm$ 0.00} & \textbf{65.02 $\pm$ 3.55} \\
        \bottomrule
    \end{tabular}
    \label{tab:real_world}
    
    \vspace{1pt}
    \parbox{\textwidth}{\raggedright Maximum linear velocity: 1.5 m/s.}
    
\end{table}

%% file: Sections/Conclusions.tex
\section{Conclusion}
\label{sec:Conclusions}
In this work, we present ADP, a novel framework that employs reinforcement learning to adaptively regulate dynamics modeling fidelity. By adjusting modeling complexity during planning, ADP addresses the fundamental trade-off between accuracy and efficiency and mitigates the feasibility limitations of traditional hierarchical methods. Comprehensive evaluations across 300 simulated BARN environments and real-world Jackal deployments demonstrate that ADP consistently enhances navigation success, safety, and efficiency relative to fixed scheduling strategies. Its integration with multiple planners, including DWA, MPPI, and Log-MPPI, underscores the generality of the framework, while superior performance on unseen BARN Challenge environments confirms its robustness and generalization capability. These findings indicate that adaptive dynamics modeling provides a powerful alternative to static parameterization, enabling navigation systems to balance computational efficiency with physical feasibility. This paradigm shift from fixed to adaptive modeling offers a promising pathway for advancing mobile robot autonomy in complex environments.

% We believe DDP represents a significant step toward bridging the gap between theoretical planning algorithms and practical mobile robot applications.
% \section*{acknowledgements}
% This work has taken place in the RobotiXX Laboratory at George Mason University. RobotiXX research is supported by National Science Foundation (NSF, 2350352), Army Research Office (ARO, W911NF2320004, W911NF2420027, W911NF2520011), Air Force Research Laboratory (AFRL), US Air Forces Central (AFCENT), Google DeepMind (GDM), Clearpath Robotics, Raytheon Technologies (RTX), Tangenta, Mason Innovation Exchange (MIX), and Walmart.

%% file: LuXiao.bbl
\begin{thebibliography}{10}
\providecommand{\url}[1]{#1}
\csname url@samestyle\endcsname
\providecommand{\newblock}{\relax}
\providecommand{\bibinfo}[2]{#2}
\providecommand{\BIBentrySTDinterwordspacing}{\spaceskip=0pt\relax}
\providecommand{\BIBentryALTinterwordstretchfactor}{4}
\providecommand{\BIBentryALTinterwordspacing}{\spaceskip=\fontdimen2\font plus
\BIBentryALTinterwordstretchfactor\fontdimen3\font minus \fontdimen4\font\relax}
\providecommand{\BIBforeignlanguage}[2]{{%
\expandafter\ifx\csname l@#1\endcsname\relax
\typeout{** WARNING: IEEEtran.bst: No hyphenation pattern has been}%
\typeout{** loaded for the language `#1'. Using the pattern for}%
\typeout{** the default language instead.}%
\else
\language=\csname l@#1\endcsname
\fi
#2}}
\providecommand{\BIBdecl}{\relax}
\BIBdecl

\bibitem{lu2025decremental}
Y.~Lu, T.~Xu, L.~Wang, N.~Hawes, and X.~Xiao, ``Decremental dynamics planning for robot navigation,'' in \emph{2025 IEEE/RSJ International Conference on Intelligent Robots and Systems (IROS)}.\hskip 1em plus 0.5em minus 0.4em\relax IEEE, 2025.

\bibitem{tang2025path}
Y.~Tang, M.~A. Zakaria, and M.~Younas, ``Path planning trends for autonomous mobile robot navigation: A review,'' \emph{Sensors}, vol.~25, no.~4, p. 1206, 2025.

\bibitem{xiao2022motion}
X.~Xiao, B.~Liu, G.~Warnell, and P.~Stone, ``Motion planning and control for mobile robot navigation using machine learning: a survey,'' \emph{Autonomous Robots}, vol.~46, no.~5, pp. 569--597, 2022.

\bibitem{khanal2023guided}
A.~Khanal, H.-D. Bui, G.~J. Stein, and E.~Plaku, ``Guided sampling-based motion planning with dynamics in unknown environments,'' in \emph{2023 IEEE 19th International Conference on Automation Science and Engineering (CASE)}.\hskip 1em plus 0.5em minus 0.4em\relax IEEE, 2023, pp. 1--8.

\bibitem{perille2020benchmarking}
D.~Perille, A.~Truong, X.~Xiao, and P.~Stone, ``Benchmarking metric ground navigation,'' in \emph{2020 IEEE International Symposium on Safety, Security, and Rescue Robotics (SSRR)}.\hskip 1em plus 0.5em minus 0.4em\relax IEEE, 2020, pp. 116--121.

\bibitem{elfes1989using}
A.~Elfes, ``Using occupancy grids for mobile robot perception and navigation,'' \emph{Computer}, vol.~22, no.~6, pp. 46--57, 1989.

\bibitem{dijkstra2022note}
E.~W. Dijkstra, ``A note on two problems in connexion with graphs,'' in \emph{Edsger Wybe Dijkstra: his life, work, and legacy}, 2022, pp. 287--290.

\bibitem{lavalle2001rapidly}
S.~M. LaValle and J.~J. Kuffner, ``Rapidly-exploring random trees: Progress and prospects: Steven m. lavalle, iowa state university, a james j. kuffner, jr., university of tokyo, tokyo, japan,'' \emph{Algorithmic and computational robotics}, pp. 303--307, 2001.

\bibitem{kavraki1996probabilistic}
L.~E. Kavraki, P.~Svestka, J.-C. Latombe, and M.~H. Overmars, ``Probabilistic roadmaps for path planning in high-dimensional configuration spaces,'' \emph{IEEE transactions on Robotics and Automation}, vol.~12, no.~4, pp. 566--580, 1996.

\bibitem{fox1997dynamic}
D.~Fox, W.~Burgard, and S.~Thrun, ``The dynamic window approach to collision avoidance,'' \emph{IEEE Robotics \& Automation Magazine}, vol.~4, no.~1, pp. 23--33, 1997.

\bibitem{fox2002dynamic}
------, ``The dynamic window approach to collision avoidance,'' \emph{IEEE robotics \& automation magazine}, vol.~4, no.~1, pp. 23--33, 2002.

\bibitem{williams2017model}
G.~Williams, A.~Aldrich, and E.~A. Theodorou, ``Model predictive path integral control: From theory to parallel computation,'' \emph{Journal of Guidance, Control, and Dynamics}, vol.~40, no.~2, pp. 344--357, 2017.

\bibitem{mohamed2022autonomous}
I.~S. Mohamed, K.~Yin, and L.~Liu, ``Autonomous navigation of agvs in unknown cluttered environments: log-mppi control strategy,'' \emph{IEEE Robotics and Automation Letters}, vol.~7, no.~4, pp. 10\,240--10\,247, 2022.

\bibitem{minavrik2024model}
M.~Mina{\v{r}}{\'\i}k, R.~P{\v{e}}ni{\v{c}}ka, V.~Von{\'a}sek, and M.~Saska, ``Model predictive path integral control for agile unmanned aerial vehicles,'' in \emph{2024 IEEE/RSJ International Conference on Intelligent Robots and Systems (IROS)}.\hskip 1em plus 0.5em minus 0.4em\relax IEEE, 2024, pp. 13\,144--13\,151.

\bibitem{pfeiffer2017perception}
M.~Pfeiffer, M.~Schaeuble, J.~Nieto, R.~Siegwart, and C.~Cadena, ``From perception to decision: A data-driven approach to end-to-end motion planning for autonomous ground robots,'' in \emph{2017 ieee international conference on robotics and automation (icra)}.\hskip 1em plus 0.5em minus 0.4em\relax IEEE, 2017, pp. 1527--1533.

\bibitem{kim2018end}
Y.-H. Kim, J.-I. Jang, and S.~Yun, ``End-to-end deep learning for autonomous navigation of mobile robot,'' in \emph{2018 IEEE International Conference on Consumer Electronics (ICCE)}.\hskip 1em plus 0.5em minus 0.4em\relax IEEE, 2018, pp. 1--6.

\bibitem{amini2019variational}
A.~Amini, G.~Rosman, S.~Karaman, and D.~Rus, ``Variational end-to-end navigation and localization,'' in \emph{2019 International Conference on Robotics and Automation (ICRA)}.\hskip 1em plus 0.5em minus 0.4em\relax IEEE, 2019, pp. 8958--8964.

\bibitem{bui2022improving}
H.-D. Bui, Y.~Lu, and E.~Plaku, ``Improving the efficiency of sampling-based motion planners via runtime predictions for motion-planning problems with dynamics,'' in \emph{2022 IEEE/RSJ International Conference on Intelligent Robots and Systems (IROS)}.\hskip 1em plus 0.5em minus 0.4em\relax IEEE, 2022, pp. 4486--4491.

\bibitem{lu2023leveraging}
Y.~Lu and E.~Plaku, ``Leveraging single-goal predictions to improve the efficiency of multi-goal motion planning with dynamics,'' in \emph{2023 IEEE/RSJ International Conference on Intelligent Robots and Systems (IROS)}.\hskip 1em plus 0.5em minus 0.4em\relax IEEE, 2023, pp. 850--857.

\bibitem{xiao2022learning}
X.~Xiao, T.~Zhang, K.~M. Choromanski, T.-W.~E. Lee, A.~Francis, J.~Varley, S.~Tu, S.~Singh, P.~Xu, F.~Xia, S.~M. Persson, L.~Takayama, R.~Frostig, J.~Tan, C.~Parada, and V.~Sindhwani, ``Learning model predictive controllers with real-time attention for real-world navigation,'' in \emph{Conference on robot learning}.\hskip 1em plus 0.5em minus 0.4em\relax PMLR, 2022.

\bibitem{xiao2022appl}
X.~Xiao, Z.~Wang, Z.~Xu, B.~Liu, G.~Warnell, G.~Dhamankar, A.~Nair, and P.~Stone, ``Appl: Adaptive planner parameter learning,'' \emph{Robotics and Autonomous Systems}, vol. 154, p. 104132, 2022.

\bibitem{wang2021apple}
Z.~Wang, X.~Xiao, G.~Warnell, and P.~Stone, ``Apple: Adaptive planner parameter learning from evaluative feedback,'' \emph{IEEE Robotics and Automation Letters}, vol.~6, no.~4, pp. 7744--7749, 2021.

\bibitem{xu2021applr}
Z.~Xu, G.~Dhamankar, A.~Nair, X.~Xiao, G.~Warnell, B.~Liu, Z.~Wang, and P.~Stone, ``Applr: Adaptive planner parameter learning from reinforcement,'' in \emph{2021 IEEE international conference on robotics and automation (ICRA)}.\hskip 1em plus 0.5em minus 0.4em\relax IEEE, 2021, pp. 6086--6092.

\bibitem{lu2025multi}
Y.~Lu, D.~Das, E.~Plaku, and X.~Xiao, ``Multi-goal motion memory,'' in \emph{2025 IEEE International Conference on Robotics and Automation (ICRA)}.\hskip 1em plus 0.5em minus 0.4em\relax IEEE, 2025.

\bibitem{das2024motion}
D.~Das, Y.~Lu, E.~Plaku, and X.~Xiao, ``Motion memory: Leveraging past experiences to accelerate future motion planning,'' in \emph{2024 IEEE International Conference on Robotics and Automation (ICRA)}.\hskip 1em plus 0.5em minus 0.4em\relax IEEE, 2024, pp. 16\,467--16\,474.

\bibitem{plaku2018clearance}
E.~Plaku, E.~Plaku, and P.~Simari, ``Clearance-driven motion planning for mobile robots with differential constraints,'' \emph{Robotica}, vol.~36, no.~7, pp. 971--993, 2018.

\bibitem{le2021multi}
D.~Le and E.~Plaku, ``Multi-robot motion planning with unlabeled goals for mobile robots with differential constraints,'' in \emph{2021 IEEE International Conference on Robotics and Automation (ICRA)}.\hskip 1em plus 0.5em minus 0.4em\relax IEEE, 2021, pp. 7950--7956.

\bibitem{khanal2024learning}
A.~Khanal, H.-D. Bui, E.~Plaku, and G.~J. Stein, ``Learning-informed long-horizon navigation under uncertainty for vehicles with dynamics,'' in \emph{2024 IEEE/RSJ International Conference on Intelligent Robots and Systems (IROS)}.\hskip 1em plus 0.5em minus 0.4em\relax IEEE, 2024, pp. 3069--3075.

\bibitem{xiao2021learning}
X.~Xiao, J.~Biswas, and P.~Stone, ``Learning inverse kinodynamics for accurate high-speed off-road navigation on unstructured terrain,'' \emph{IEEE Robotics and Automation Letters}, vol.~6, no.~3, pp. 6054--6060, 2021.

\bibitem{karnan2022vi}
H.~Karnan, K.~S. Sikand, P.~Atreya, S.~Rabiee, X.~Xiao, G.~Warnell, P.~Stone, and J.~Biswas, ``Vi-ikd: High-speed accurate off-road navigation using learned visual-inertial inverse kinodynamics,'' in \emph{2022 IEEE/RSJ International Conference on Intelligent Robots and Systems (IROS)}.\hskip 1em plus 0.5em minus 0.4em\relax IEEE, 2022, pp. 3294--3301.

\bibitem{pokhrel2024cahsor}
A.~Pokhrel, M.~Nazeri, A.~Datar, and X.~Xiao, ``{CAHSOR}: Competence-aware high-speed off-road ground navigation in $\mathbb{SE}(3)$,'' \emph{IEEE Robotics and Automation Letters}, 2024.

\bibitem{datar2024learning}
A.~Datar, C.~Pan, and X.~Xiao, ``Learning to model and plan for wheeled mobility on vertically challenging terrain,'' \emph{IEEE Robotics and Automation Letters}, 2024.

\bibitem{datar2024terrain}
A.~Datar, C.~Pan, M.~Nazeri, A.~Pokhrel, and X.~Xiao, ``Terrain-attentive learning for efficient 6-dof kinodynamic modeling on vertically challenging terrain,'' in \emph{2024 IEEE/RSJ International Conference on Intelligent Robots and Systems (IROS)}.\hskip 1em plus 0.5em minus 0.4em\relax IEEE, 2024, pp. 5438--5443.

\bibitem{fujimoto2018addressing}
S.~Fujimoto, H.~Hoof, and D.~Meger, ``Addressing function approximation error in actor-critic methods,'' in \emph{International conference on machine learning}.\hskip 1em plus 0.5em minus 0.4em\relax PMLR, 2018, pp. 1587--1596.

\bibitem{hasselt2010double}
H.~Hasselt, ``Double q-learning,'' \emph{Advances in neural information processing systems}, vol.~23, 2010.

\bibitem{xiao2022autonomous}
X.~Xiao, Z.~Xu, Z.~Wang, Y.~Song, G.~Warnell, P.~Stone, T.~Zhang, S.~Ravi, G.~Wang, H.~Karnan \emph{et~al.}, ``Autonomous ground navigation in highly constrained spaces: Lessons learned from the benchmark autonomous robot navigation challenge at icra 2022 [competitions],'' \emph{IEEE Robotics \& Automation Magazine}, vol.~29, no.~4, pp. 148--156, 2022.

\bibitem{xiao2023autonomous}
X.~Xiao, Z.~Xu, G.~Warnell, P.~Stone, F.~G. Guinjoan, R.~T. Rodrigues, H.~Bruyninckx, H.~Mandala, G.~Christmann, J.~L. Blanco-Claraco \emph{et~al.}, ``Autonomous ground navigation in highly constrained spaces: Lessons learned from the second barn challenge at icra 2023 [competitions],'' \emph{IEEE Robotics \& Automation Magazine}, vol.~30, no.~4, pp. 91--97, 2023.

\bibitem{xiao2024autonomous}
X.~Xiao, Z.~Xu, A.~Datar, G.~Warnell, P.~Stone, J.~J. Damanik, J.~Jung, C.~A. Deresa, T.~D. Huy, C.~Jinyu \emph{et~al.}, ``Autonomous ground navigation in highly constrained spaces: Lessons learned from the third barn challenge at icra 2024 [competitions],'' \emph{IEEE Robotics \& Automation Magazine}, vol.~31, no.~3, pp. 197--204, 2024.

\bibitem{xiaoautonomous}
X.~Xiao, Z.~Xu, S.~A. Ghani, A.~Datar, D.~Song, P.~Stone, A.~Mazen, K.~Yazdipaz, I.~Mateyaunga, M.~Faied \emph{et~al.}, ``Autonomous ground navigation in highly constrained spaces: Lessons learned from the forth barn challenge at icra 2025.''

\bibitem{damanik2024lics}
J.~J. Damanik, J.-W. Jung, C.~A. Deresa, and H.-L. Choi, ``Lics: Navigation using learned-imitation on cluttered space,'' \emph{IEEE Robotics and Automation Letters}, 2024.

\end{thebibliography}
